\documentclass{article}

\usepackage[final, nonatbib]{neuripsDeepInverse_2020}

\usepackage[utf8]{inputenc} % allow utf-8 input
\usepackage[T1]{fontenc}    % use 8-bit T1 fonts
\usepackage{hyperref}       % hyperlinks
\usepackage{url}            % simple URL typesetting
\usepackage{booktabs}       % professional-quality tables
\usepackage{amsfonts}       % blackboard math symbols
\usepackage{nicefrac}       % compact symbols for 1/2, etc.
\usepackage{microtype}      % microtypography
\usepackage{graphicx}
\usepackage{colortbl}
\usepackage{color}
\usepackage{float}
\usepackage{xifthen}
\usepackage{hyperref}       % hyperlinks
\usepackage{url}            % simple URL typesetting
\usepackage{booktabs}       % professional-quality tables
\usepackage{amsfonts}       % blackboard math symbols
\usepackage{nicefrac}       % compact symbols for 1/2, etc.
\usepackage{microtype}      % microtypography
\usepackage{amsmath}
\usepackage{xcolor}
\usepackage{soul}
\usepackage{enumitem}
\usepackage{svg}

\definecolor{amber}{rgb}{1, 0.75, 0} 

\newcommand{\round}[1]{\left( #1 \right)}

\newcommand{\curly}[1]{\left\{ #1 \right\}}

\newcommand{\CQ}{{\mathcal{Q}}}
\newcommand{\bbR}{{\mathbb{R}}}
\newcommand{\CC}{{\mathcal{C}}}
\newcommand{\CL}{{\mathcal{L}}}
\newcommand{\CD}{{\mathcal{D}}}

\newcommand{\CR}{{\mathcal{R}}}
\newcommand{\CN}{{\mathcal{N}}}
\newcommand{\KL}{{\mathrm{KL}}}

\newcommand{\ELBO}{\mathrm{ELBO}}
\newcommand{\NLL}{\mathrm{NLL}}
\newcommand{\relu}{\mathrm{relu}}
\newcommand{\diag}{\mathrm{diag}}
\newcommand{\Var}{\mathrm{Var}}

\newcommand{\qps}[1]{q_{\ifthenelse{\isempty{#1}}{\psi}{\psi_{#1}}}}
\newcommand{\qPs}[1]{q_{\ifthenelse{\isempty{#1}}{\Psi}{\Psi_{#1}}}}
\newcommand{\qpS}[1]{q_{\ifthenelse{\isempty{#1}}{\Psi}{\Psi_{#1}}}}
\newcommand{\qpsp}[1]{q_{\ifthenelse{\isempty{#1}}{\psi}{\psi_{#1}},\Phi_{k}}}
\newcommand{\qpSp}[1]{q_{\ifthenelse{\isempty{#1}}{\Psi}{\Psi_{#1}},\Phi_{k}}}
\newcommand{\qpsa}[1]{q^\ast_{\ifthenelse{\isempty{#1}}{\psi}{\psi_{#1}}}}
\newcommand{\qpSa}[1]{q^\ast_{\ifthenelse{\isempty{#1}}{\Psi}{\Psi_{#1}}}}
\newcommand{\qpsap}[1]{q^\ast_{\ifthenelse{\isempty{#1}}{\psi}{\psi_{#1}},\Phi_{k}}}
\newcommand{\qpSap}[1]{q^\ast_{\ifthenelse{\isempty{#1}}{\Psi}{\Psi_{#1}},\Phi_{k}}}

\providecommand{\argmin}{\operatorname*{argmin}}

\newcommand\tab[1][0.cm]{\hspace*{#1}}

\usepackage[linesnumbered,ruled,vlined]{algorithm2e}
\SetCommentSty{mycommfont}
\SetKwInput{KwInput}{Input}                % Set the Input
\SetKwInput{KwOutput}{Output}              % set the Output

\newcommand{\removed}[1]{}

\title{Quantifying Sources of Uncertainty in \\{Deep Learning-Based} Image Reconstruction}

\author{
  Riccardo Barbano\thanks{Corresponding author.} \\
  University College London, UK\\
  \texttt{riccardo.barbano.19@ucl.ac.uk} \\
   \And
   {\v Z}eljko Kereta \\
   University College London, UK\\
   \texttt{z.kereta@ucl.ac.uk} \\
   \And
   Chen Zhang \\
   Huawei Technologies R\&D UK\\
   \texttt{chenzhang10@huawei.com}
   \\
   \And
   Andreas Hauptmann\\
   University of Oulu, Finland \\
   University College London, UK\\
   \texttt{andreas.hauptmann@oulu.fi} \\
   \And
   Simon Arridge\\
   University College London, UK\\
   \texttt{s.arridge@ucl.ac.uk} \\
  \And
   Bangti Jin\\
   University College London, UK\\
   \texttt{b.jin@ucl.ac.uk} \\
}

\begin{document}

\maketitle

\begin{abstract}
Image reconstruction methods based on deep neural networks have shown outstanding performance, equalling or exceeding the state-of-the-art results of conventional approaches, but often do not provide uncertainty information about the reconstruction. In this work we propose a scalable and efficient framework to simultaneously quantify aleatoric and epistemic uncertainties in learned iterative image reconstruction. We build on a Bayesian deep gradient descent method for quantifying epistemic uncertainty, and incorporate the heteroscedastic variance of the noise to account for the aleatoric uncertainty. 
We show that our method exhibits competitive performance against conventional benchmarks for computed tomography with both sparse view and limited angle data. The estimated uncertainty captures the variability in the reconstructions, caused by the restricted measurement model, and by missing information, due to the limited angle geometry.
\end{abstract}

\section{Introduction}\label{sec:intro}
% auto-ignore 
% !TEX root = main.tex
In the past few years deep learning (DL) based image reconstruction techniques have demonstrated remarkable empirical results. 
As examples in medical imaging, we mention approaches that replace components of established optimisation algorithms (e.g., gradient descent \cite{putzky2017recurrent, adler2017solving, hauptmann2018model}, primal-dual algorithm \cite{adler2018learned}, ADMM \cite{sun2016deep}) by DNNs.
See \cite{arridge2019solving,ongie2020deep} for overviews. The overwhelming majority of these techniques are deterministic, and there is a lack of solutions that provide mechanisms to estimate the uncertainty.

The absence of quantitative estimates of predictive uncertainty, bias, and robustness has greatly hindered the applicability of many deep learning methods in sensitive domains \cite{gal2016uncertainty,Kendall:2017}, such as medical imaging and autonomous driving. There are several types of uncertainty in the context of deep learning \cite{Kendall:2017, hullermeier2019aleatoric}. These include aleatoric and epistemic uncertainties, which originate from the stochastic variability inherent in the data generating process, and from the uncertainty in the parameters of the model, respectively. Uncertainty quantification can be conveniently formulated within a Bayesian framework. Although easy to formulate, such frameworks can be challenging to use since computing quantities from the posterior is computationally intractable. In addition, learning-based methods that incorporate uncertainty quantification often do not outperform, but rather are, at best, on par with classical non-Bayesian approaches \cite{Osawa:2019}. 

There have been several recent attempts to characterise uncertainty in deep learning-based methods for solving inverse problems, see \cite{zhang2019probabilistic,adler2018deep} for aleatoric uncertainty and \cite{Barbano:2020} for epistemic uncertainty, but the topic is still largely in its infancy. In \cite{Barbano:2020} a greedy, iterative, data-driven, knowledge-aided, Bayesian approach, termed Bayesian deep gradient descent (BDGD), is developed to solve inverse problems in imaging. In this work we extend BDGD so that aleatoric and epistemic uncertainties can both be quantified. 

We develop a hybrid architecture (in an unrolled optimisation scheme) by appending a Gaussian mean-field CNN to a deterministic CNN-based feature extractor. Moreover, epistemic and aleatoric uncertainties are accounted for using a decomposition approach proposed in \cite{depeweg2018decomposition}, which is achieved by bifurcating the network's architecture. Due to its hybrid nature, the resulting Bayesian framework is easy to train using stochastic variational inference (VI) \cite{rezende2014stochastic}, and allows us to improve the reconstruction quality over deterministic DNN techniques. 
Our main contributions can be summarised as follows. Firstly, we introduce a statistically principled framework to quantify aleatoric and epistemic uncertainties in image reconstruction by integrating a data-driven, knowledge-aided framework with the advances in BNNs and VI. Secondly, we apply the framework to computed tomography (CT) reconstructions, where our results yield interpretable uncertainty. 

% \AH{Depending on which workshop we submit, we need to be careful with saying CT, as only have simulated parallel beam experiments. It might be better to say X-ray tomography}
\section{Model Specification}\label{sec:model}
% auto-ignore 
% !TEX root = main.tex
\begin{figure}[!ht]
\begin{minipage}{0.48\textwidth}
In image reconstruction we aim to recover an image $x$ from (noisy) measurements $y$, given a forward operator $A$, such that $Ax=y$. An established way to obtain a reconstruction is by solving the variational problem 
\begin{equation}
\label{eqn:regularisation}
    x^{\ast} \in \argmin_{x\in\CC} \left\{ \CL_{\mathrm{MAP}} = \CD(y, Ax) + \lambda \CR(x)\right\},
\end{equation}
where $\CD(y, Ax)$ is the data fidelity and $\CR(x)$ a regularising penalty term, $\mathcal{C}$ a constraint set, and $\lambda \in \mathbb{R}^{+}$ balances the two terms. The solution $x^*$ is often found by iterative optimisation methods (with initial guess $x_0$).
This produces a sequence of iterates, and the iterations terminate once a stopping criterion is satisfied.
\end{minipage}\,\,\,
\begin{minipage}{0.50\textwidth}
\centering
\includegraphics[width=1\textwidth]{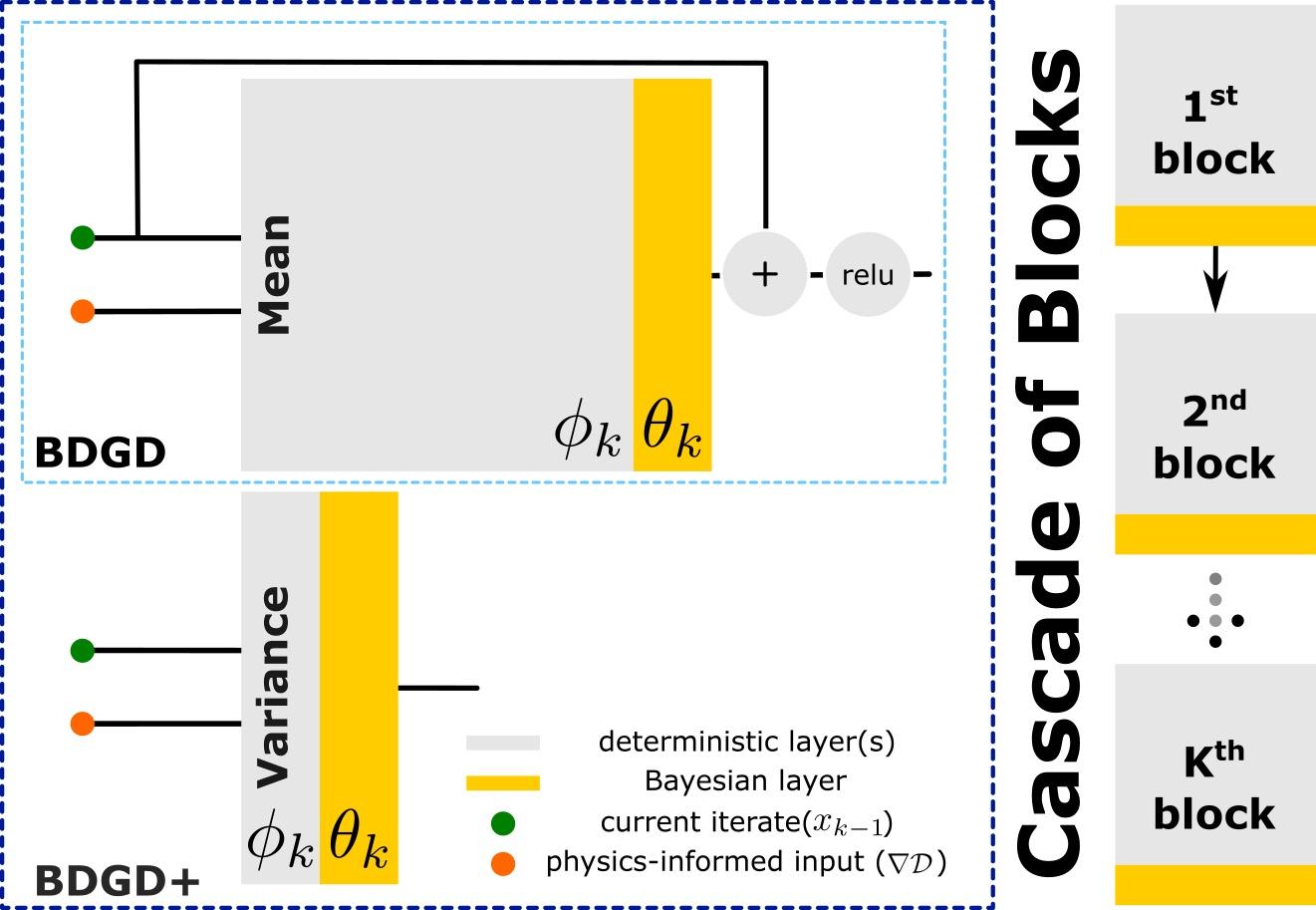}
\vspace{-0.5cm}
\caption{A diagram of the overall BDGD/BDGD+ framework. Further details are in Appendix \ref{appx:AdditionalArchFigures}.}
\label{fig:architecture}
\end{minipage}
\end{figure}
% To recover a (vectorised)\AH{I would not vectorise it, we also do not seem to need the dimensions $L$ later} image $x \in \bbR^L$ from (noisy) measurements $y\in \bbR^m$, standard image reconstruction methods solve a regularised minimisation problem 
% \begin{equation}
%     x^{\ast} \in \argmin_{x\in\CC} \left\{ \CL_{\mathrm{MAP}} = \CD(y, Ax) + \lambda \CR(x)\right\},
% \end{equation}
% where $\CD(y, Ax)$ is the data fidelity, $A$ the forward operator, $\CR(x)$ a penalty, $\mathcal{C}$ a constraint set, and $\lambda \in \mathbb{R}^{+}$ balances the two terms. The solution $x^*$ is often found by iterative methods (with initial guess $x_0$). This produces a sequence of iterates, and it is stopped once a stopping criteria is satisfied.

\noindent\textbf{Unrolled optimisation} \cite{GregorLecun:2010,putzky2017recurrent} is a prominent DL paradigm for image reconstruction. It mimics an iterative method, but executes only a fixed number of iterations, and at each iteration the update is computed using a DNN. In this work we consider the iterates computed as residual update by
% \AH{Doesn't skip often concatenate? I probably would just say, as "...iterates computed as residual update by"}
\[
x_{k} = \relu(x_{k-1} + \delta x_{k-1}),
\]
where the rectifier linear unit, $\relu(x):=\max(0, x)$, and the increment $\delta x_{k-1}$ is obtained by
\begin{equation}
\delta x_{k-1} = f_{\phi_{k}}\left(\nabla \CD\left(y, A x_{k-1}\right), x_{k-1}\right)=:f_{\phi_k}(\nabla\CD, x_{k-1}).
\end{equation}
To simplify the notation we overload the mappings $f_{\phi_k}$ and write the iterates as $x_{k}=f_{\phi_k}(\nabla\CD,x_{k-1})$. We refer to mappings (networks) $f_{\phi_{k}}$ as \emph{blocks}, and the collection of all the blocks defining the iterations, as a \emph{cascade}.
In a cascade $K$ blocks are applied sequentially, and the reconstructed image for an initial guess $x_0$ is given by
\begin{equation}\label{eqn:cascade}
x_{K} = \left( f_{\phi_{K}}\circ f_{\phi_{K-1}} \circ 
    \cdots \circ f_{\phi_{1}}\right) (\nabla \CD, x_{0}) := f_{\Phi_{K}}(\nabla \CD, x_{0}), \,\, \text{ with } \Phi_k:=(\phi_1,\ldots,\phi_k).
\end{equation}
Fig. \ref{fig:architecture} shows the overall framework. Note that $x_k$ is at each iteration  progressively refined with the information passed through the gradient $\nabla \CD\left(y, Ax_{k-1}\right)$, thus incorporating the physical model underlying the inverse problem into the reconstruction process. 

\noindent\textbf{Specifics of the model and its training.} In our architecture each block consists of deterministic layers and one (mean-field CNN) Bayesian layer (see Fig. \ref{fig:architecture}). In the following, we will omit the deterministic parameters from the discussion, for the sake of clarity of notation. Within a Bayesian framework the network parameters $\theta_k$ are random variables, and we approximate the posterior with VI. In the $k^{\text{th}}$ block we compute an optimal approximate distribution $\qpsa{k}(\theta)$, parametrised by $\psi_k$, such that it approximates the true posterior $p(\theta|X,Y)$, where $X,Y$ is our dataset consisting of ground truth data and corresponding observations. Moreover, we train the cascade greedily, i.e.,  one block at a time.
Let $\Theta_k:=(\theta_1,\ldots,\theta_k)$. 
Assume the first $k-1$ blocks have been learnt, i.e., $\qpSa{k-1}$ is known. We consider the variational family $\mathcal{Q}_k$ that consists of distributions of the form \[\qpS{k}(\Theta_k)=\qpSa{k-1}(\Theta_{k-1})\qps{k}(\theta_k;\Theta_{k-1}),\text{ with }\qps{k}(\theta_k;\Theta_{k-1}) = \prod^{D}_{d=1} \mathcal{N}(\mu_{k, d}, \sigma_{k, d}^2),\]
where $\psi_k=\{(\mu_{k,\ell}, \sigma_{k, d}^2)\}^{D}_{d=1}$ are the variational parameters, and $D$ is the number of parameters per block. In the $k^{\text{th}}$ block, we optimise for
\begin{equation}\label{eqn:qpSak}
    \begin{split}
	\qpSa{k}\!:=\qpSa{k}(\cdot;X,Y)\in\!\argmin_{\qpS{k}\in\CQ_k}\mathcal{L}_k(\qpS{k}; X, Y), 
% 	=\qpSa{k-1}(\Theta_{k-1})\qpsa{k}(\theta_k;\Theta_{k-1}),\text{ with } \qpsa{k}(\theta_k;\Theta_{k-1})
	\end{split}
\end{equation}
% where $\CQ$ is a family of $D$-dimensional Gaussians, 
where the loss function $\mathcal{L}_k$ is given by
\begin{equation}\label{eqn:Li}
    \mathcal{L}_k(\qpS{k}; X, Y)\! :=\!\!-\int\!\! \qpS{k}(\Theta_k)\log p(X|Y,\Theta_k)\mathrm{d}{\Theta_k}\!+\! \KL(\qpS{k}(\Theta_k)\Vert p(\Theta_k)).
\end{equation}
The prior and the likelihood remain to be specified. The former is set recursively as 
\[
p(\Theta_{k})=\qpSa{k-1}(\Theta_{k-1})p(\theta_k;\Theta_{k-1}), \quad \text{ where } p(\theta_k;\Theta_{k-1})=\CN(0,I).
\]
Choosing the likelihood adequately allows the capture of either aleatoric or epistemic uncertainty, or both. 
% In our architecture each block consists of deterministic layers and one (mean-field CNN) Bayesian layer (see Fig. \ref{fig:architecture}), whose distributional parameters $\psi_k$ are learnt.
To do this, in BDGD \cite{Barbano:2020} we take the likelihood as
\begin{equation}\label{eqn:exp_likelihood}p(x|y,\Theta_{k})=\mathcal{N}(f_{\Theta_k}(\nabla \CD, x_{0}),\sigma_k^2 I).\end{equation} 
%we capture parametric uncertainty. 
Note that we model isotropic homoscedastic noise with variance $\sigma^2_k$ (a trainable parameter). 
Following \cite{Kendall:2017}, to capture aleatoric uncertainty, we can use input-dependent variance, and set the likelihood to be 
\begin{equation}\label{eqn:var_likelihood}p(x|y, \Theta_{k}) = \mathcal{N}\left(f_{\Theta_{k}}(\nabla \CD, x_{0}), \diag( \sigma^{2}_{\Theta_{k}}(\nabla \CD, x_{0}))\right).\end{equation}
Note that the network $\sigma^{2}_{\Theta_{k}}(\nabla \CD, x_{0})$ is an \emph{a posteriori} heteroscedastic model. We refer to this model as BDGD+. Once the model is trained, following \cite{depeweg2018decomposition}, we can decouple aleatoric and epistemic uncertainties by decomposing the (entry-wise) predictive variance $\Var[x]$ at the $K^{\rm th}$ step, using the law of total variance, and estimating with $T\geq 1$ Monte Carlo samples as 
\begin{align}\label{eqn:uncertainty_decomposition}
\begin{split}
    \Var[x] &= \Var_{\qpS{K}(\Theta_K)}[\mathbb{E}(x|y,\Theta_K)] + \mathbb{E}_{\qPs{K}(\Theta_K)}[\Var(x|y, \Theta_K)]\\ &\approx \underbrace{\dfrac{1}{T} \sum_{t=1}^{T} \sigma_{\Theta^t_K}^2(\nabla\CD, x_0)}_{\text{aleatoric}} + 
    \underbrace{\dfrac{1}{T} \sum_{t=1}^{T} f_{\Theta_K^t}(\nabla\CD,x_0)^{2}-\left(\dfrac{1}{T}\sum_{t=1}^{T}f_{\Theta_K^t}(\nabla\CD,x_0)\right)^{2}}_{\text{epistemic}}.
\end{split}
\end{align}
All the operations in \eqref{eqn:uncertainty_decomposition} are meant entry-wise, and $\Theta_{K}^t \sim\qpSa{K}(\Theta_{K})$ for each $t=1,\ldots, T$. We denote the Monte-Carlo estimate on the right-hand side, as $\widehat\Var[x]$. More details and practicalities regarding the training of our hybrid architecture, and the inference, can be found in Appendix \ref{appx:PractInTraining}.

\section{Experimental Results}
% auto-ignore 
% !TEX root = main.tex
We showcase the performance of BDGD+ for CT reconstructions. Here the forward map $A$ is given by the discrete Radon transform and we can compute a direct reconstruction by filtered back-projection (FBP).
As the reconstruction problem is ill-posed, FBP reconstructions typically exhibit strong artefacts, especially if only a subset of data is available. 
%This can be represented by a composition $A = S\circ R$, where $S$ is a sub-sampling of the directions. 
In the following, we study two cases of practical interest: sparse view and limited angle. In the former, $\ell$ directions (here $\ell \in \{8, 16, 32, 64, 128\}$) are taken uniformly from $0^\circ$ to $180^\circ$; in the latter, the maximum range of available angle data is less than $180^\circ$. We add $1\%$ Gaussian noise to the sinogram and the data fidelity is the squared $L^2$ norm $\frac{1}{2}\|Ax-y\|^2_2$, so that
\begin{equation}
    \nabla \CD(y, Ax_{k-1}) = A^{\top}(Ax_{k-1} -y).
\end{equation}
%where $A^{\top}$ denotes the (unfiltered) back-projection operator. 

The depth $K$ of the cascade in BDGD+ (and BDGD and DGD) depends on the problem difficulty.
For sparse view we use $K=20$ for 8 and 16 directions (further increasing $K$ does not lead to better reconstructions), and $K=10$ for 32, 64, and 128 directions. For limited angle we use $K=30$ for $[0^\circ, 90^\circ)$ and $[0^\circ, 120^\circ)$, and $K=20$ for $[0^\circ, 150^\circ)$. The initial guess $x_{0}$ is given by the FBP reconstruction, and we train each block for 150 epochs. All the methods are trained on 4000 randomly generated ellipses (all of size $128\times128$), and are tested on 100 ellipses, and on the Shepp-Logan phantom. Further details, including the benchmarks used, can be found in Appendix \ref{appx:BenchImplementationDetails}. 

\begin{figure}[H]
    \centering
    %trim=0mm 0mm 10mm 0mm, clip
    \includegraphics[width=1\textwidth]{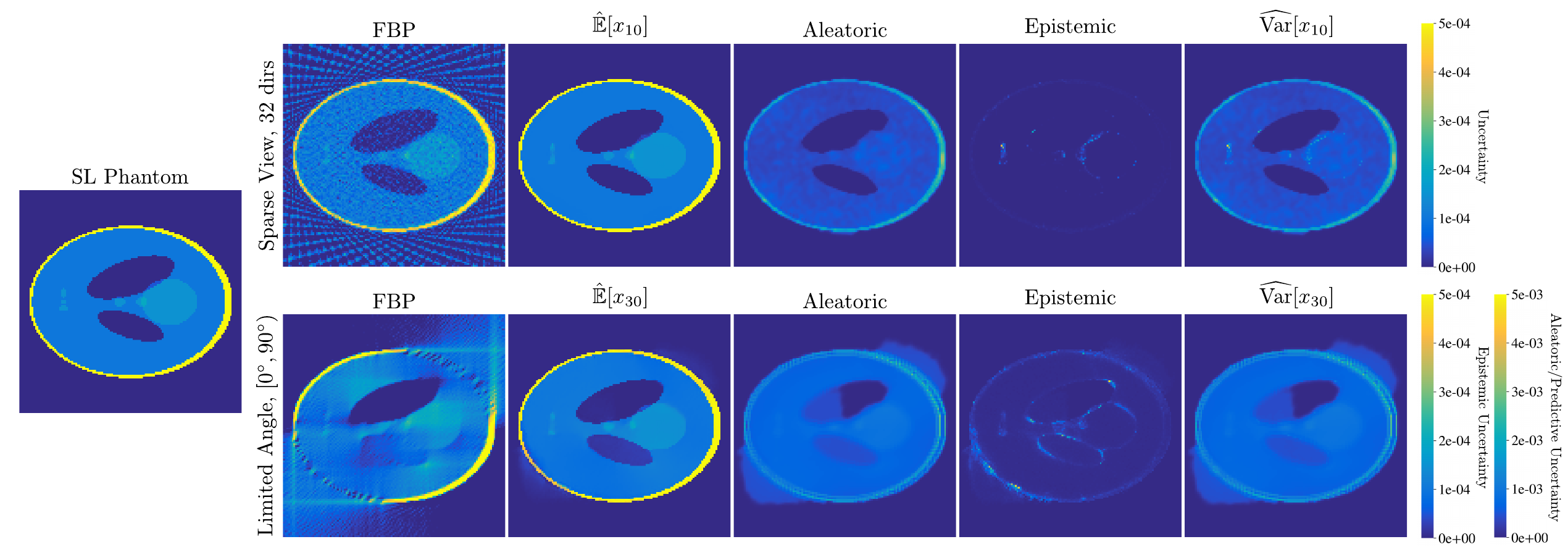}
    \caption{The reconstructions for sparse view CT with 32 directions (top) and limited angle with $[0,90^\circ)$ (bottom); see Appendix \ref{appx:AdditionalExpFigures} for further relevant results.}
    \label{fig:SparseViewMainUncertain}
\end{figure}

The results are presented in Table \ref{table:SparseandLimitedViewCT} and Fig. \ref{fig:SparseViewMainUncertain}. The values in Table \ref{table:SparseandLimitedViewCT} denote the peak signal to noise ratio (PSNR) of the reconstruction (i.e., the mean for BDGD and BDGD+). Clearly, BDGD+ is competitive with benchmarks used:  FBP+U-Net \cite{ronneberger2015u}, LPD \cite{adler2018learned} and DGD \cite{hauptmann2018model},
which are all far more accurate than non-deep learning based techniques, i.e., FBP and TV \cite{chambolle2011tv} (where the regulariser, in \eqref{eqn:regularisation}, is the total variation, $\CR(x)=\|\nabla x\|_1$). Interestingly, BDGD+ consistently improves on BDGD in terms of PSNR, and the improvement can be nontrivial. This shows the importance of modelling heteroscedastic variance in obtaining high-quality reconstructions. 

\begin{table}[H]
\centering
\caption{Sparse view, with respect to the number of directions (dirs), and limited angle, with respect to the available range of angles. 
The first reported number is the mean PSNR over the ellipses, and the second number is the PSNR for the Shepp-Logan.}
\resizebox{.95\textwidth}{!}{%
\begin{tabular}{cccccccccc}
&\multicolumn{5}{c}{\textsf{\textbf{Sparse View}}} & & \multicolumn{3}{c}{\textsf{ \textbf{Limited Angle}}}\\
\cmidrule{2-6} \cmidrule{8-10}
\multicolumn{1}{c}{\textsf{\textbf{{Methods}}}}     & 8 dirs (95$\%$ red.) & 16 dirs (91$\%$ red.) & 32 dirs (82$\%$ red.) & 64 dirs (64$\%$ red.) & 128 dirs (29$\%$ red.)  && $[0^\circ, 90^\circ)$ & $[0^\circ, 120^\circ)$ & $[0^\circ, 150^\circ)$\\ \cmidrule{1-6} \cmidrule{8-10}
\cmidrule{1-6} \cmidrule{8-10}
\multicolumn{1}{c}{\textsf{\textbf{{{FBP}}}}}  & 16.08/10.09 & 20.30/14.08 & 24.86/18.96 & 29.11/23.75 & 31.85/25.82 & &13.75/14.23 & 17.28/17.11 &  22.87/20.19 \\
\multicolumn{1}{c}{\textsf{\textbf{{{TV}}}}} & 28.33/17.90   &   32.11/35.51    &  34.93/35.63             & 35.80/36.19    &  36.54/36.47 && 28.00/26.87  & 31.15/29.31 &  34.21/33.59 \\
\multicolumn{1}{c}{\textsf{\textbf{FBP}} \textsf{\textbf{+}} \textsf{\textbf{U-Net}}} &  28.22/19.20 &  33.44/25.37 & 39.10/31.57 &  44.47/ 41.87 & 48.18/46.47 &&  13.73/14.22  & 37.78/28.21 & 42.80/35.47  \\
\multicolumn{1}{c}{\textsf{\textbf{LPD}}}  & 30.71/23.21 & \textbf{38.97}/\textbf{37.90} & \textbf{44.73}/43.09 & 47.94/48.37 & 49.42/47.15 && \textbf{35.96}/30.57  & 39.75/30.94 & 45.37/41.26 \\
\multicolumn{1}{c}{\textsf{\textbf{DGD}}}  & \textbf{31.64}/\textbf{24.17} & 38.40/\textbf{39.97} & 43.40/\textbf{45.63} & 47.27/49.03 & 50.45/51.35 && \textbf{35.56}/\textbf{35.83} & \textbf{39.88}/\textbf{42.12} & 45.25/47.52 \\
\rowcolor{amber}
\multicolumn{1}{c}{\textsf{\textbf{BDGD}}}  & 30.04/21.35 & 37.08/37.32 & 42.30/41.88 & \textbf{48.06}/\textbf{50.64} & \textbf{51.85}/\textbf{54.39} && 32.18/29.67 &  37.49/36.81 & \textbf{45.91}/\textbf{49.55} \\
\rowcolor{amber}
\multicolumn{1}{c}{\textsf{\textbf{BDGD+}}} & \textbf{31.33}/\textbf{23.82} & \textbf{38.92}/37.39 & \textbf{45.01}/\textbf{45.08} & \textbf{48.86}/\textbf{51.65} & \textbf{53.00}/\textbf{56.89} && 33.73/\textbf{32.81} & \textbf{40.60}/\textbf{44.45} &\textbf{48.78/52.23}\\ \hline
\end{tabular}%
}
\label{table:SparseandLimitedViewCT}
\end{table}

Decomposition \eqref{eqn:uncertainty_decomposition} allows us to separately quantify aleatoric and epistemic uncertainties. In both sparse view and limited angle reconstructions, aleatoric uncertainty appears to dominate, with its overall shape close to the mean (but of a smaller magnitude). In contrast, epistemic uncertainty is localised to certain regions (and is of a smaller magnitude), capturing the "out of distribution" (not in the training data) text in sparse view (cf. Fig. \ref{fig:SparseViewMain} in the Appendix), or capturing artefacts due to limited angle data. In particular, we stress that in the latter case, aleatoric uncertainty captures the limits of the geometry.
Thus, aleatoric and epistemic uncertainties provide complementary information about the reconstructions, and might shed different insights into their reliability.

% \begin{figure}[H]
%     \centering
%     %trim=0mm 0mm 10mm 0mm, clip
%     \includegraphics[width=0.85\textwidth]{figures/sparse view uncertainty 2.png}
%     \caption{Sparse view CT with 32 directions, cf. Figs. \ref{fig:DeepSparseView}, \ref{fig:SparseViewMain} in Appendix \ref{appx:AdditionalExpFigures}.}
%     \label{fig:SparseViewMainUncertain}
% \end{figure}
% \vspace{-1.5em}
% \begin{figure}[H]
%     \centering
%     % trim=0mm 0mm 22mm 0mm, clip
%     \includegraphics[width=0.80\textwidth]{figures/limited view uncertainty.png}
%     \caption{LAT for $[0^\circ, 90^\circ)$, cf. Fig. \ref{fig:DeepLimitedView} in Appendix \ref{appx:AdditionalExpFigures}. 
%     %\AH{We should add the ground-truth phantom here as well}
%     }
%     \label{fig:LimitedViewMain}
% \end{figure} 

\section{Conclusion}
% auto-ignore 
% !TEX root = main.tex
In this work we design a hybrid framework for quantifying the uncertainty in image reconstruction for inverse problems. The experimental results with sparse view/limited angle CT show that BDGD+ is competitive with conventional benchmarks, and can improve reconstruction quality in terms of PSNR over BDGD. The uncertainty maps can capture variabilities in the reconstructions caused by different sources, e.g., limited information or out of distribution data.

\bibliography{reference.bib}

\begin{thebibliography}{10}

\bibitem{putzky2017recurrent}
P.~Putzky and M.~Welling, ``Recurrent inference machines for solving inverse
  problems,'' {\em arXiv:1706.04008}, 2017.

\bibitem{adler2017solving}
J.~Adler and O.~{\"O}ktem, ``Solving ill-posed inverse problems using iterative
  deep neural networks,'' {\em Inverse Problems}, vol.~33, no.~12, p.~124007,
  2017.

\bibitem{hauptmann2018model}
A.~Hauptmann, F.~Lucka, M.~Betcke, N.~Huynh, J.~Adler, B.~Cox, P.~Beard,
  S.~Ourselin, and S.~Arridge, ``Model-based learning for accelerated,
  limited-view 3-d photoacoustic tomography,'' {\em IEEE Trans. Med. Imag.},
  vol.~37, no.~6, pp.~1382--1393, 2018.

\bibitem{adler2018learned}
J.~Adler and O.~{\"O}ktem, ``Learned primal-dual reconstruction,'' {\em IEEE
  Trans. Med. Imag.}, vol.~37, no.~6, pp.~1322--1332, 2018.

\bibitem{sun2016deep}
J.~Sun, H.~Li, Z.~Xu, {\em et~al.}, ``Deep {ADMM}-{N}et for compressive sensing
  {MRI},'' in {\em NIPS}, pp.~10--18, 2016.

\bibitem{arridge2019solving}
S.~Arridge, P.~Maass, O.~{\"O}ktem, and C.-B. Sch{\"o}nlieb, ``Solving inverse
  problems using data-driven models,'' {\em Acta Numerica}, vol.~28,
  pp.~1--174, 2019.

\bibitem{ongie2020deep}
G.~Ongie, A.~Jalal, R.~G. Baraniuk, C.~A. Metzler, A.~G. Dimakis, and
  R.~Willett, ``Deep learning techniques for inverse problems in imaging,''
  {\em IEEE J. Sel. Areas Inf. Theory}, pp.~39 -- 56 in press, 2020.

\bibitem{gal2016uncertainty}
Y.~Gal, {\em Uncertainty in {D}eep {L}earning}.
\newblock PhD thesis, University of Cambridge, 2016.

\bibitem{Kendall:2017}
A.~Kendall and Y.~Gal, ``What uncertainties do we need in {B}ayesian deep
  learning for computer vision?,'' in {\em Proceedings of the 31st
  International Conference on Neural Information Processing Systems},
  pp.~5580--5590, 2017.

\bibitem{hullermeier2019aleatoric}
E.~H{\"u}llermeier and W.~Waegeman, ``Aleatoric and epistemic uncertainty in
  machine learning: A tutorial introduction,'' {\em arXiv preprint
  arXiv:1910.09457}, 2019.

\bibitem{Osawa:2019}
K.~Osawa, S.~Swaroop, M.~E.~E. Khan, A.~Jain, R.~Eschenhagen, R.~E. Turner, and
  R.~Yokota, ``Practical deep learning with {B}ayesian principles,'' in {\em
  NIPS}, 2019.

\bibitem{zhang2019probabilistic}
C.~Zhang and B.~Jin, ``Probabilistic residual learning for aleatoric
  uncertainty in image restoration,'' {\em arXiv:1908.01010}, 2019.

\bibitem{adler2018deep}
J.~Adler and O.~{\"O}ktem, ``Deep {B}ayesian inversion,'' {\em arXiv preprint
  arXiv:1811.05910}, 2018.

\bibitem{Barbano:2020}
R.~Barbano, C.~Zhang, S.~Arridge, and B.~Jin, ``Quantifying model-uncertainty
  in inverse problems via {B}ayesian deep gradient descent.'' Preprint,
  arXiv:2007.09971, 2020.

\bibitem{depeweg2018decomposition}
S.~Depeweg, J.-M. Hernandez-Lobato, F.~Doshi-Velez, and S.~Udluft,
  ``Decomposition of uncertainty in {B}ayesian deep learning for efficient and
  risk-sensitive learning,'' in {\em International Conference on Machine
  Learning}, pp.~1184--1193, 2018.

\bibitem{rezende2014stochastic}
D.~J. Rezende, S.~Mohamed, and D.~Wierstra, ``Stochastic backpropagation and
  approximate inference in deep generative models,'' {\em arXiv preprint
  arXiv:1401.4082}, 2014.

\bibitem{GregorLecun:2010}
K.~Gregor and Y.~{LeCun}, ``Learning fast approximations of sparse coding,'' in
  {\em ICML}, pp.~1--8, 2010.

\bibitem{ronneberger2015u}
O.~Ronneberger, P.~Fischer, and T.~Brox, ``U-net: Convolutional networks for
  biomedical image segmentation,'' in {\em International Conference on Medical
  image computing and computer-assisted intervention}, pp.~234--241, Springer,
  2015.

\bibitem{chambolle2011tv}
A.~Chambolle and T.~Pock, ``A first-order primal-dual algorithm for convex
  problems with applications to imaging,'' {\em Journal of Mathematical Imaging
  and Vision}, no.~40, pp.~120--145, 2011.

\bibitem{RudinOsherFatemi:1992}
L.~I. Rudin, S.~Osher, and E.~Fatemi, ``Nonlinear total variation based noise
  removal algorithms,'' {\em Phys. D}, vol.~60, no.~1-4, pp.~259--268, 1992.

\bibitem{hauptmann2020unreasonable}
A.~Hauptmann and J.~Adler, ``On the unreasonable effectiveness of cnns,'' {\em
  arXiv preprint arXiv:2007.14745}, 2020.

\bibitem{adler2017odl}
J.~Adler, H.~Kohr, and O.~Oktem, ``Operator discretization library (odl),''
  {\em Software available from https://github.com/odlgroup/odl}, 2017.

\bibitem{van2016fast}
W.~Van~Aarle, W.~J. Palenstijn, J.~Cant, E.~Janssens, F.~Bleichrodt,
  A.~Dabravolski, J.~De~Beenhouwer, K.~J. Batenburg, and J.~Sijbers, ``Fast and
  flexible x-ray tomography using the {ASTRA} toolbox,'' {\em Optics Expr.},
  vol.~24, no.~22, pp.~25129--25147, 2016.

\end{thebibliography}

\clearpage
\section*{Appendices}
% auto-ignore 
% !TEX root = main.tex
\appendix

\section{Additional Materials for Bayesian Deep Gradient Descent (BDGD)}\label{appx: }

In this part, we provide further details about the proposed framework. In Appendix \ref{appx:PractInTraining}, we discuss the training and inference of BDGD and BDGD+. In Appendix \ref{appx:AleatoricUncertainty}, we review how we model aleatoric and epistemic uncertainties. In \ref{appx:2BlockCascade} we analyse a 2-block cascade, for the sake of clarity. 

\subsection{Practicalities in Training and Inference}\label{appx:PractInTraining}

Hereafter, we denote the deterministic parameters of the (deterministic) feature extractor as $\phi_k$, the random variable of the mean-field CNN layer as $\theta_{k}$, and the variational parameters defining the Bayesian layer as $\psi_k$. In the proposed framework, we use a composition of maps $f_{\theta_k}\circ f_{\phi_k}$ to model the $k^{\text{th}}$ block, and perform VI on the parameters $\theta_k$'s in a greedy manner.

During the training we optimise jointly with respect to $\phi_k$'s as well. In doing so, $\phi_k$'s are in effect regarded as hyperparameters. Hence, in the $k^{\text{th}}$ block (by abusing the notation) we solve for
\begin{equation}
\begin{split}
	\min_{\qpSp{k}\in\CQ_k,\phi_k} \{\mathcal{L}(\phi_{k}, \qpSp{k}; X, Y) &:= -\int \qpSp{k} (\Theta_k)\log p_{\Phi_{k}}(X|Y,\Theta_k)\mathrm{d}{\Theta_k}\\ &\hspace{1em} + \mathrm{KL}(\qpSp{k}(\Theta_k)||p_{\Phi_{k}}(\Theta_k))\}, 
\end{split}
\end{equation}
where $p_{\Phi_{k}}(X|Y,\Theta_k)$ and $\qpSp{k}(\Theta_k)$ are respectively the likelihood and the approximate posterior distribution parametrised by the hyperparameters $\Phi_{k}$. 
%Again, for notational simplicity, in this section we omit $\Phi$ in the notation. 
%For instance, we denote $(f_{\theta_k}\circ f_{\phi_k}\circ \cdots \circ f_{\theta_1}\circ f_{\phi_1})(\nabla \CD,\cdot)=:f_{\Theta_k}(\nabla \CD, \cdot)$, instead of $f_{ \Phi_{k}, \Theta_{k}}(\nabla \CD, \cdot)$, where $x_k=f_{ \Phi_{k}, \Theta_{k}}(\nabla \CD, x_0)$. 
Below we suppress the deterministic parameters $\Phi$ from notation, and focus only on the probabilistic ones.
Methodologically, this is equivalent to setting the variational family to be a delta approximation, i.e., mean field with zero variance, on some parameters, but a Gaussian mean field approximation on the remaining ones. Note that we can add an $L^{2}$ penalty on deterministic parameters $\Phi_{K}$ to the lower bound functional. This does not change the statistical interpretation, since it does not penalise the random parameter $\theta$, and we can add the penalty right after the KL divergence.

The hybrid approach greatly reduces the number of variational parameters, especially if the Bayesian component is only a small portion of the overall network. In doing so, the resulting cascade has an overall complexity comparable with its deterministic counterpart, while retaining the ability to quantify uncertainty. 

Once all the blocks in the cascade are trained, it can be used for inference. Each sampling step amounts to a feed forward propagation through the network, which is computationally very efficient (at least when compared with classical iterative reconstruction algorithms). Recall and the likelihood for BDGD is
\begin{equation*}
p(x|y,\Theta_{K}) = \mathcal{N}(f_{\Theta_{K}}(\nabla D, x_0),\sigma^2_{K}I),
\end{equation*}
and for BDGD+ is 
\begin{equation*} p(x|y, \Theta_{k}) = \mathcal{N}\left(f_{\Theta_{k}}(\nabla \CD, x_{0}), \diag( \sigma^{2}_{\Theta_{k}}(\nabla \CD, x_{0}))\right).
\end{equation*}
The approximate posterior distribution is given by 
\begin{equation}\qpSa{K}(\Theta_{K})=\qpsa{1}(\theta_1)\prod_{k=2}^{K} \qpsa{k}(\theta_k;\Theta_{k-1}).
\end{equation} At test time, for each input $y$, we can use MC to estimate the statistics, i.e., the mean and the covariance, of the approximate predictive distribution
\begin{equation}
    \qpSa{K}(x|y) = \int p(x|y,\Theta_{K})\qpSa{K}(\Theta_{K}){\rm{d}}{\Theta_{K}}.
\end{equation}
Specifically, $\mathbb{E}[x]$ can be estimated with the unbiased empirical estimator
\begin{equation}
    \hat{\mathbb{E}}[x]:=\dfrac{1}{T}\sum_{t=1}^{T}f_{\hat{\Theta}^{t}_{K}}(\nabla \CD, x_0)\xrightarrow[\;\;T\;\;\xrightarrow{}\;\;\infty\;\;]{}
    \mathbb{E}[x], 
\end{equation}
with $T$ samples of $\hat{\Theta}_K$ from $\qpSa{K}(\Theta_{K})$, i.e., $\{\hat{\Theta}^{t}_{K}\}_{t=1}^T$. Moreover, the predictive uncertainty of $\qpSa{K}(x|y)$ can be estimated with the sample covariance matrix
\begin{equation}\label{eqn:predictive_covariance_matrix}
   \widehat{\mathrm{Cov}}[x] := \sigma^{2}_K I + \dfrac{1}{T}\sum_{t=1}^{T} f_{\hat{\Theta}^{t}_{K}}(\nabla \CD, x_0)^{\otimes 2} - \big(\dfrac{1}{T}\sum_{t=1}^{T} f_{\hat{\Theta}^{t}_{K}}(\nabla \CD, x_0)\big)^{\otimes 2},
\end{equation}
where $x^{\otimes 2} = xx^{\top}$. Indeed,
\begin{align*}
    \mathbb{E}[x]
    &=\int x\qpsa{K}(x|y)\mathrm{d}x
    = \int \int x\mathcal{N}(f_{\Theta_K}(\nabla \CD, x_0), \sigma^{2}_K I)\qpSa{K}(\Theta_K) \mathrm{d}\Theta_K\mathrm{d}x\\
    &= \int \left(\int x\mathcal{N}(f_{\Theta_K}(\nabla \CD, x_0), \sigma_{K}^{2}I) \mathrm{d}x\right)\qpSa{K}(\Theta_K)\mathrm{d}\Theta_K
    = \int f_{\Theta_K}(\nabla \CD, x_0)\qpSa{K}(\Theta_K)\mathrm{d}\Theta_K.
\end{align*}
And
\begin{align*}
    \mathbb{E}\left[x^{\otimes 2}\right]
    &=\int\left(\int x^{\otimes 2} \mathcal{N}(f_{\Theta_K}(\nabla \CD, x_0), \sigma^{2}_K I)\mathrm{d} x\right) \qpSa{K}(\Theta_K) \mathrm{d}\Theta_K\\ 
    &=\int\left(\operatorname{Cov}_{p\left(x | y, \Theta_K\right)}\left[x\right]+\mathbb{E}_{p\left(x | y, \Theta_K\right)}\left[x\right]^{\otimes 2}\right)\qpSa{K}(\Theta_K) \mathrm{d}\Theta_K\\
    &=\!\int\!\big(\sigma^{2}_K I\!+\!f_{\Theta_K}(\nabla \CD, x_0)^{\otimes 2}\big) \qpSa{K}(\Theta_K) \mathrm{d}\Theta_K\!=\!\sigma^{2}_K I \!+ \!\int\! f_{\Theta_K}(\nabla \CD, x_0)^{\otimes 2} \qpSa{K}(\Theta_K) \mathrm{d}\Theta_K.
\end{align*}

Note that the above derivations suggest an isotropic homoscedastic noise model. 

It follows that $\hat{\mathbb{E}}[x]$ and $\widehat{\mathrm{Cov}}[x]$ are unbiased MC estimators of $\mathbb{E}[x]$ and $\mathrm{Cov}[x]=\mathbb{E}\left[x^{\otimes 2}\right]-\mathbb{E}[x]^{\otimes 2}$ with $T$ samples. Note that we use the diagonal element of the sample covariance matrix to quantify the corresponding predictive uncertainty. 
The training and inference procedures of our proposed framework are summarised by Algorithms 1 and 2, respectively.

After training all the blocks, we reconstruct the next update $x_{k}$ with 1 MC sample, and compute the gradient of the data fidelity term $\CD$, which is problem specific, dependent on the forward operator and noise statistics. At inference we use 100 MC samples to estimate the mean image and the pixel-wise variance $\widehat{\mathrm{Var}}[x]$, which is a vector consisting of the diagonal elements in $\widehat{\mathrm{Cov}}[x]$.

\subsection{Modelling Aleatoric Uncertainty via a Heteroscedastic Noise Model}\label{appx:AleatoricUncertainty}

The framework in \cite{Barbano:2020} can be viewed as an example of an isotropic homoscedastic, block-wise noise model, as the variance is not a function of the input, and it is fixed at all spatial locations, i.e., $\sigma^2_{k}I$ is background noise. This is a restrictive and unrealistic assumption for most imaging modalities. 

In this work we introduce an anisotropic heteroscedastic noise model, i.e., BDGD+. For computational tractability, heteroscedastic anisotropic modelling assumes that the covariance matrix is diagonal, i.e., $\diag(\sigma^{2}_{\Theta_{k}}(\nabla\CD, x_{0}))$. 

We model the likelihood $p(x|y, \Theta_{k})$ of the $k^{\text{th}}$ block as a Gaussian distribution with input-varying variance,
\begin{align*}
    p(x|y, \Theta_{k}) &= \mathcal{N}\left(f_{\Theta_{k}}(\nabla \CD, x_{0}), \mathrm{diag}(\sigma^{2}_{\Theta_{k}}(\nabla\CD, x_{0}))\right),\\
    &= \dfrac{\exp\left(-\dfrac{1}{2}\left(x - f_{\Theta_{k}}(\nabla \CD, x_{0})\right)^{\top}\diag(\sigma^{2}_{\Theta_{k}}(\nabla\CD, x_{0}))^{-1} \left(x - f_{\Theta_{k}}(\nabla \CD, x_{0})\right)\right)}{(2\pi)^{\frac{L}{2}} \sqrt{\prod_{\ell=1}^{L} [\sigma^2_{\Theta_{k}}(\nabla\CD, x_{0})}]_{\ell}},
\end{align*}
where $f_{\Theta_{k}}(\nabla \CD, x_{0})$ is the mean (the reconstructed image), $\diag(\sigma^{2}_{\Theta_{k}}(\nabla\CD, x_{0}))$ is the covariance matrix, and thus a function of the input $x_{0}$, $[\cdot]_{\ell}$ denotes the $\ell^{\rm th}$ entry of a vector, and $L$ is the number of pixels (the dimensionality) of $x$. 

When jointly optimising the loss for the $k^{\text{th}}$ block, the negative log-likelihood, $\NLL(\qpS{k}; X,Y)$, over a mini-batch set $\mathcal{B}$ is defined as 
\begin{align}
    \NLL(\qpS{k}; X, Y) &= \sum_{i\in \mathcal{B}} -\log \mathcal{N}\left(f_{\Theta_{k}}(\nabla \CD, x_{i, 0}), \diag(\sigma^{2}_{\Theta_{k}}(\nabla\CD, x_{i, 0}))\right)\\
        &= \mathcal{M}(\qpS{k}; X, Y) + \mathcal{H}(\qpS{k}; X, Y) + c, 
\end{align}
where $c$ is an absolute constant, and $x_{i,0}$ is the FBP of the data sample $x_i$. 
The first term $\mathcal{M}(\qpS{k}; X, Y)$ is the squared Mahalanobis distance over the mini-batch, defined as 
\begin{equation}
    \mathcal{M}(\qpS{k}; X, Y)=\frac{1}{|\mathcal{B}|} \sum_{i\in\mathcal{B}}\left(x_{i} - f_{\Theta_{k}}(\nabla \CD, x_{i, 0})\right)^{\top} \diag(\sigma^{2}_{\Theta_{k}}(\nabla\CD, x_{i, 0})^{-1}
    \left(x_{i} - f_{\Theta_{k}}(\nabla \CD, x_{i, 0})\right).
\end{equation}
The second term $\mathcal{H}(\qpS{k}; X, Y)$ is the mean differential entropy over the mini-batch $B$
\begin{equation}
    \mathcal{H}(\qpS{k}; X, Y)=\frac{1}{|\mathcal{B}|} \sum_{i\in\mathcal{B}} \sum_{\ell=1}^{L} \log[\sigma^{2}_{\Theta_{k}}(\nabla\CD, x_{i, 0})]_{\ell}.
\end{equation}

When the covariance is diagonal, the first term $\mathcal{M}(\qpS{k}; X, Y)$ corresponds to the mean squared error weighted by the inverse of the corresponding variance. The second term $\mathcal{H}(\qpS{k}; X, Y)$ prevents the variance from growing too large.
 \begin{algorithm}[H]
    \caption{Training}
    \KwInput{\small number of reconstruction steps $K$, dataset $X,Y$, initial guesses $x_{i,0}$, batch-size $|\mathcal{B}|$\\
    Compute FBPs $x_{i,0}$ of all data samples $x_i$}
    \For{$k$ $\leftarrow$ $\mathrm{1}$ $\mathrm{to\;} K$}{
        Construct the block's input:\\
        $\tab[1cm]\mathcal{D}_{k-1} = \{x_{i, k-1}, \nabla \CD(y_{i}, Ax_{i, k-1})\}^{N}_{i=1}$\\
        Train the $k^{\text{th}}$ block $f_{\phi_{k}, \theta_{k}}(\nabla \CD(y_{i}, Ax_{i, k-1}), x_{i, k-1})$:\\
        \CommentSty{$\tab[0.3cm]$// stochastic mini-batch optimisation}\\
        $\tab[0.3cm] \psi^{\ast}_{k}, \phi^{\ast}_{k}\leftarrow \argmin_{\qpSp{k} \in \CQ_k, \phi_{k}}\bigg\{\hat{\mathcal{L}}(\phi_{k}, \qpSp{k}; \mathcal{D}_{k-1}):= \tab[0.8cm]-\dfrac{N}{|\mathcal{B}|}\displaystyle\sum_{i\in \mathcal{B}} 
        \int \qpSp{k}(\Theta_k) \log p_{\Phi_{k}}(x_{i}|y_{i},\Theta_k) \mathrm{d}\Theta_{k} + \mathrm{KL}(\qpSp{k}(\Theta_k)||p_{\Phi_{k}}(\Theta_k))\bigg\}$\\
        \CommentSty{// update with $\hat{\theta}_k\sim \qpsap{k}(\theta_k;\Theta_{k-1})$}\\
        $x_{i, k}\leftarrow f_{\phi_{k},{\hat{\theta}}_{k}}(\nabla \CD(y_{i}, Ax_{i, k-1}), x_{i, k-1})$
        }
    \KwOutput{\small approximate posterior at each reconstruction step}
    \label{algo:1}
  \end{algorithm}

 \begin{algorithm}[H]
    \caption{Inference}
    \KwInput{\small observation $y$, trained cascade parameters ($\Phi_K, \Psi_K$), the number $T$ of MC samples}
    \CommentSty{//we set T to 100\\}
    \For{$t$ $\leftarrow$ $\mathrm{1}$ $\mathrm{to\;} T$}{
        \CommentSty{//with $\hat{\Theta}_K^{t}\sim q^{\ast}_{\Psi_{K},\Phi_{K}}(\Theta_K)$}\\
        Sample $x^{t}_K = f_{\Phi_K, \hat{\Theta}_K^{t}}(\nabla \CD(y, Ax_0), x_0)$\\
        }
    Evaluate $\hat{\mathbb{E}}[x]$ and $\widehat{\mathrm{Var}}[x]$ with $\{x^{t}_K\}_{t=1}^T$\\
    \KwOutput{\small $\hat{\mathbb{E}}[x]$ and $\widehat{\mathrm{Var}}[x]$}
    \label{algo:2}
  \end{algorithm}

Thus, we quantify the uncertainty of the reconstructed signal at the $k^{\text{th}}$ block by computing the predictive uncertainty as
\begin{equation}
   \widehat{\mathrm{Cov}}[x] =  \dfrac{1}{T}\sum_{t=1}^{T} \diag(\sigma^{2}_{\hat{\Theta}^t_{k}}(\nabla\CD, x_{0})) +  \dfrac{1}{T}\sum_{t=1}^{T} f_{\hat{\Theta}^{t}_{K}}(\nabla \CD, x_0)^{\otimes 2} -\big(\dfrac{1}{T}\sum_{t=1}^{T} f_{\hat{\Theta}^{t}_{K}}(\nabla \CD, x_0)\big)^{\otimes 2}.
\end{equation}
with $\Theta_{K}^t \sim\qpSa{K}(\Theta_{K})$.
As a reminder, note that $\widehat\Var[x]$, in \eqref{eqn:uncertainty_decomposition}, is the diagonal of $\widehat{\mathrm{Cov}}[x]$.

\subsection{Example: 2-block Cascade}\label{appx:2BlockCascade}

We will now analyse a fully-Bayesian 2-block cascade to explain our methodology. 
For additional clarity we set $D=1$. 
Let $\theta_1$ and $\theta_2$ be the parameters of the first and second blocks, respectively, whereas $\psi_1$ and $\psi_2$ the variational parameters of the first and second blocks, respectively. 
As a reminder, through VI we aim to approximate the true posterior distribution $p(\theta_1,\theta_2|X,Y)$ with a simpler distribution $\qpS{2}(\theta_1,\theta_2)$.
When training the first block, we minimise the following functional with respect to the variational parameters

\begin{equation}
	\CL_1(\qps{1}; X,Y) = -\int \qps{1}(\theta_1)\log p(X|Y,\theta_1){\rm{d}}\theta_1 + \KL(\qps{1}(\theta_1)||p(\theta_1)),
\end{equation}
where $p(\theta_1)=\frac{1}{2}\exp(-\theta_1^2/2)$ is the prior distribution of $\theta_1$, and $\qps{1}(\theta_1)=\frac{1}{2\sigma_1^2}\exp\round{-\frac{(\theta_1-\mu_1)^2}{2\sigma_1^2}}$, with $\psi_1=(\mu_1, \sigma_1^2)$, is the approximate distribution in the mean field Gaussian family. Moreover, with a sample of $\theta_1$, our first block $f_{\theta_1}(\nabla \CD, x_{0})$ outputs the mean of $p(x|y,\theta_1)$. Minimising $\CL_1(\qps{1}; X,Y)$, for given data $X, Y$, we obtain an (approximate) optimal posterior distribution of $\theta_1$
\[\qpsa{1}= \argmin_{\qps{1}\in\CQ_1}\CL_1(\qps{1}; X,Y).\] 

This is used to construct the (joint) prior distribution of $(\theta_1, \theta_2)$, which is defined by $p(\theta_1, \theta_2)=\qpsa{1}(\theta_1)p(\theta_2;\theta_1)$, where $p(\theta_2;\theta_{1})=\frac{1}{2}\exp(-\theta_2^2/2)$ is the standard Gaussian distribution. 
On the other hand, the approximate posterior distribution of $(\theta_1, \theta_2)$ is pursued among distributions of the form $\qpS{2}(\theta_1,\theta_2)=\qpsa{1}(\theta_1)\qps{2}(\theta_2;\theta_1)$, where 
$\qps{2}(\theta_2;\theta_1)=\frac{1}{2\sigma_2^2}\exp\round{-\frac{(\theta_2-\mu_2)^2}{2\sigma_2^2}}$, and where $\psi_2=(\mu_2, \sigma_2^2)$.
In other words,
\begin{equation}
    \begin{split}
	\CL_2(\qpS{2}; X, Y) = &-\int \qpsa{1}(\theta_1) \qps{2}(\theta_2;\theta_1)\log p(X|Y,\theta_1, \theta_2){\rm{d}\theta_1\rm{d}\theta_2} \\&+ \mathbb{E}_{\qpsa{1}(\theta_1)}[\KL(\qps{2}(\theta_2;\theta_1)||p(\theta_2;\theta_1))].
	\end{split}
	\label{eq:lossk2}
\end{equation}
Alternatively, we can say that the approximately optimal joint distribution is computed as
\begin{equation}
	\qpSa{2}(\theta_1,\theta_2) := \qpSa{2}(\theta_1,\theta_2; X, Y,\psi_1)= \argmin_{\qpS{2}\in\CQ_2}\KL(\qpS{2}(\theta_1,\theta_2)||p(\theta_1,\theta_2|X, Y)),
\end{equation}
over the constrained variational family
\begin{equation}
	\CQ_2 = \curly{\qpS{2}(\theta_1,\theta_2)|\qpS{2}(\theta_1,\theta_2)=\qpsa{1}(\theta_1)\qps{2}(\theta_2;\theta_1)}\simeq \curly{(\mu_2,\sigma_{2}^{2})\vert \mu_2\in\bbR, \sigma^{2}_{2}\in\bbR^{\ge0}}.
\end{equation}
Note that during training  for samples $\theta_1 \sim \qpsa{1}(\theta_1)$ and $\theta_2 \sim \qpsa{2}(\theta_{2}; \theta_1)$, the composite function $f_{\theta_2}(f_{\theta_1}(\nabla \CD, x_{0}))$ is outputting the mean of $p(x|y,\theta_1, \theta_2)$. 

The derivation of the loss in \eqref{eq:lossk2} follows as 
\begin{align*}
    \KL(\qpS{2}(\theta_1, \theta_2)||p(\theta_1, \theta_2 |X, Y))
    &= \int \qpS{2}(\theta_1, \theta_2) \mathrm{log}\;\dfrac{\qpS{2}(\theta_1, \theta_2)}{p(\theta_1, \theta_2|X, Y)}\mathrm{d}\theta_1 \mathrm{d}\theta_2\\
    &= \int \qpS{2}(\theta_1, \theta_2) \mathrm{log}\;\dfrac{\qpS{2}(\theta_1, \theta_2)}{p(\theta_1, \theta_2, X|Y)}\mathrm{d}\theta_1 \mathrm{d}\theta_2 + \mathrm{log}\;p(X|Y)\\
    &\propto \int \qpS{2}(\theta_1, \theta_2) \mathrm{log}\;\dfrac{\qpS{2}(\theta_1, \theta_2)}{p(\theta_1, \theta_2, X|Y)}\mathrm{d}\theta_1 \mathrm{d}\theta_2 = - \ELBO(\qpS{2}; X, Y).
\end{align*}
Let us now expand the negative Evidence Lower BOund $\ELBO(\qpS{2}; X, Y)$ term, 
\begin{align*}
    \mathcal{L}_{2}&(\qpS{2}; X, Y) = - \int \qpS{2}(\theta_1, \theta_2)\mathrm{log}\;p(X|\theta_1, \theta_2, Y)\mathrm{d}\theta_1\mathrm{d}\theta_2 + \int \qpS{2}(\theta_1, \theta_2) \mathrm{log}\;\dfrac{\qpS{2}(\theta_1, \theta_2)}{p(\theta_1, \theta_2)}\mathrm{d}\theta_1 \mathrm{d}\theta_2\\
    &=\!-\!\int\! \qpsa{1}(\theta_1)\qps{2}(\theta_2;\theta_1)\mathrm{log}\;p(X|\theta_1, \theta_2, Y)\mathrm{d}\theta_1\mathrm{d}\theta_2\! +\!\int\! \qpsa{1}(\theta_1)\qps{2}(\theta_2;\theta_1) \mathrm{log}\;\dfrac{\qps{2}(\theta_2;\theta_1)}{p(\theta_2;\theta_1)}\mathrm{d}\theta_1\mathrm{d}\theta_2.
\end{align*}
Note that this can be interpreted as $\KL_\star(\qpS{k}(\theta_k; \Theta_{k-1})\Vert p(\Theta_k|X,Y))$ with respect to a weighted measure $(\qpSa{k}(\Theta_{k-1})d\Theta_{k-1} d\theta_k)$. 
\vfill
\section{Benchmarks Implementation Details}\label{appx:BenchImplementationDetails}

We consider four benchmark approaches:
\begin{itemize}
\item Total variation regularisation (TV) \cite{RudinOsherFatemi:1992};
\item FBP + U-Net \cite{ronneberger2015u};
\item Deep gradient descent (DGD) \cite{hauptmann2018model};
\item Learned primal-dual (LPD) \cite{adler2018learned}. 
\end{itemize}
Total variation \cite{RudinOsherFatemi:1992} is an established image reconstruction technique, suitable for recovering piecewise constant like images.
The latter three are well-established deep unrolled iteration approaches. Specifically, in FBP + U-Net, U-Net learns the identity map, as it fails to recover the missing information in the limited angle setting. The latter could be overcome by either adding an additional loss term \cite{hauptmann2018model} or using extensive training data \cite{hauptmann2020unreasonable}. DGD can be considered a greedy version of LGS \cite{adler2017solving}, and it is used here for consistency with BDGD.

TV reconstruction is computed with the Chambolle-Pock algorithm, with the regularisation parameter selected via grid search. 
BDGD/BDGD+ and the benchmarks are all implemented in Python using the Operator Discretisation Library (ODL) \cite{adler2017odl}, PyTorch and TensorFlow. To evaluate the operator $A$ and its adjoint, we use the GPU accelerated ASTRA backend \cite{van2016fast}.  For LPD we set the common parameters of training, e.g., number of iterations/epochs, batch size and number of batches, the same as DGD/BDGD to enforce comparability. Note that LPD is an unrolled scheme trained end-to-end.
Hence, the parameters of the model are trained simultaneously instead of greedily.

\section{Additional Experimental Figures}\label{appx:AdditionalExpFigures}

\begin{figure}[ht]
    \centering
    \includegraphics[width=0.9\textwidth
    % ,trim=0mm 0mm 15mm 0mm, clip
    ]{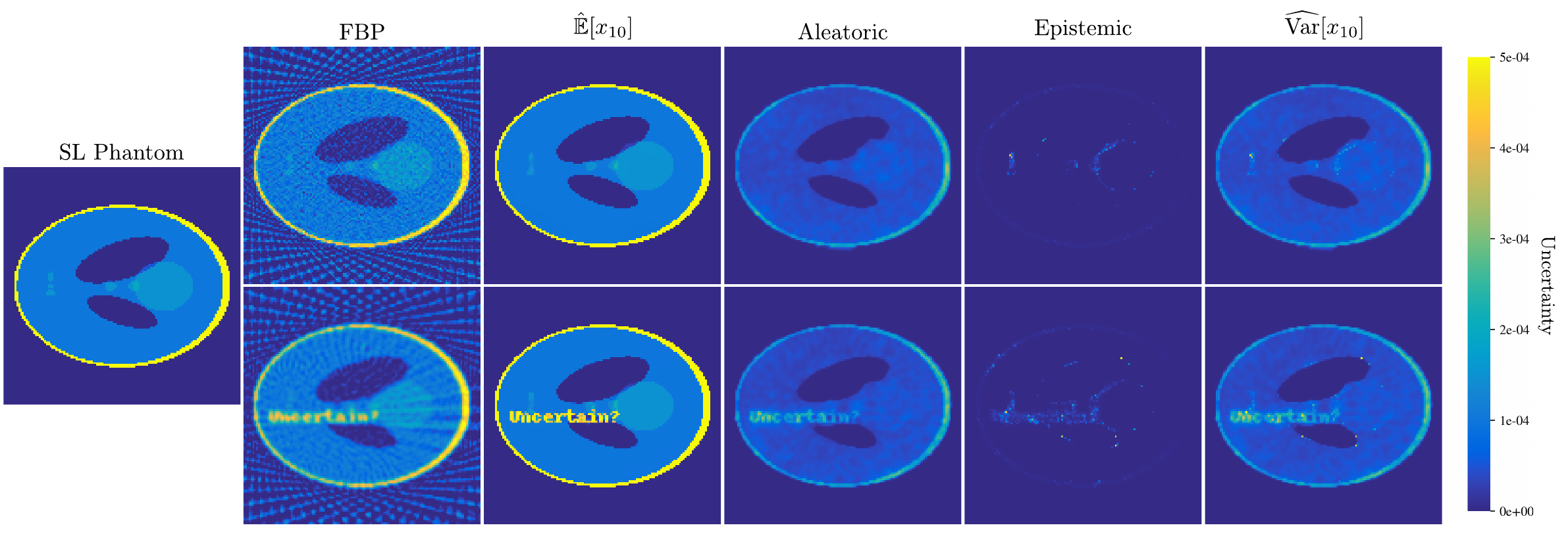}
    \caption{Sparse view CT with 32 directions.}
    \label{fig:SparseViewMain}
\end{figure}

\begin{figure}[ht]
    \centering
    \includegraphics[width=0.9\textwidth]{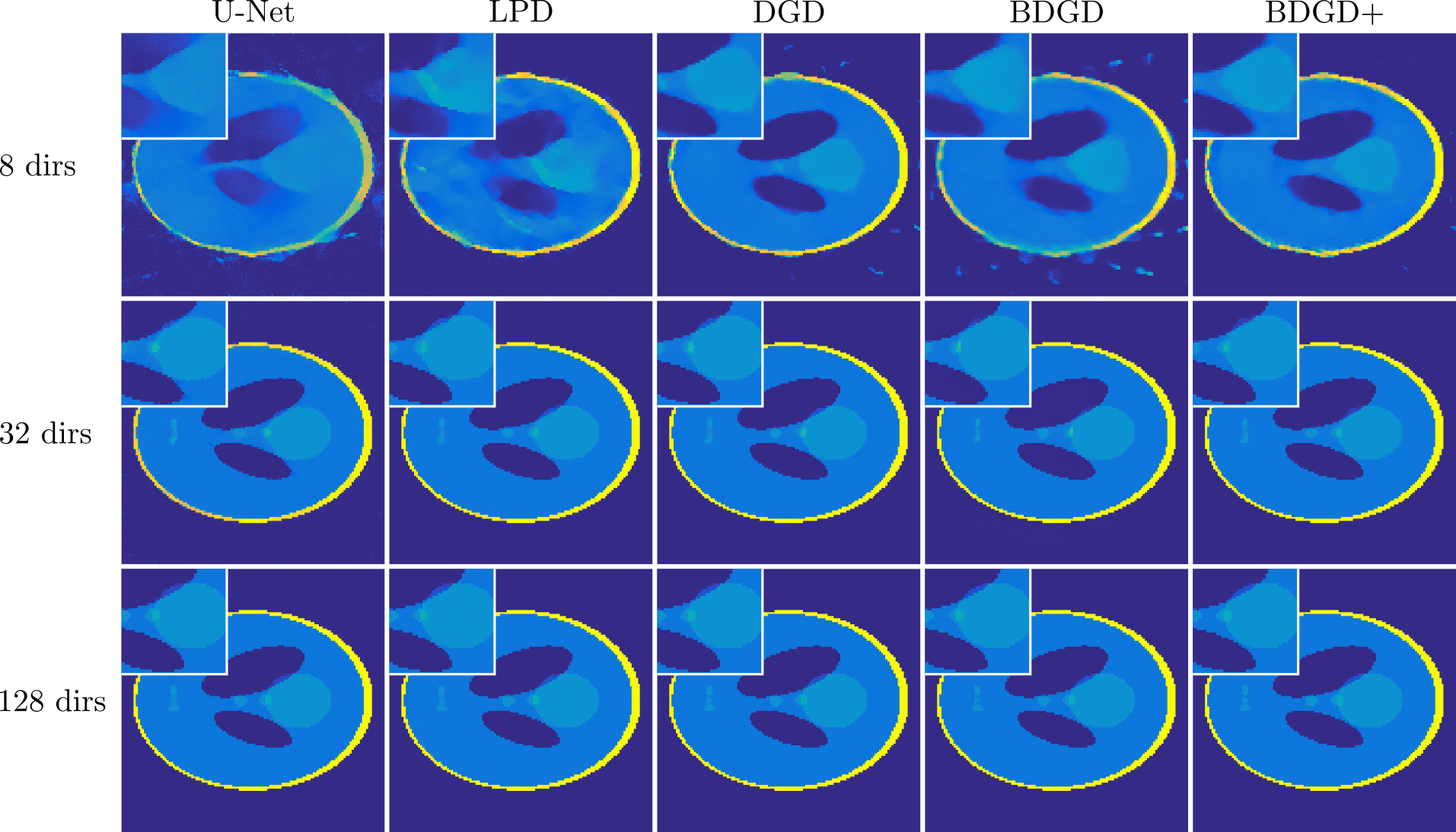}
    \caption{Deep learning approaches for sparse view CT.}
    \label{fig:DeepSparseView}
\end{figure}

\begin{figure}[ht]
    \centering
    \includegraphics[width=0.9\textwidth]{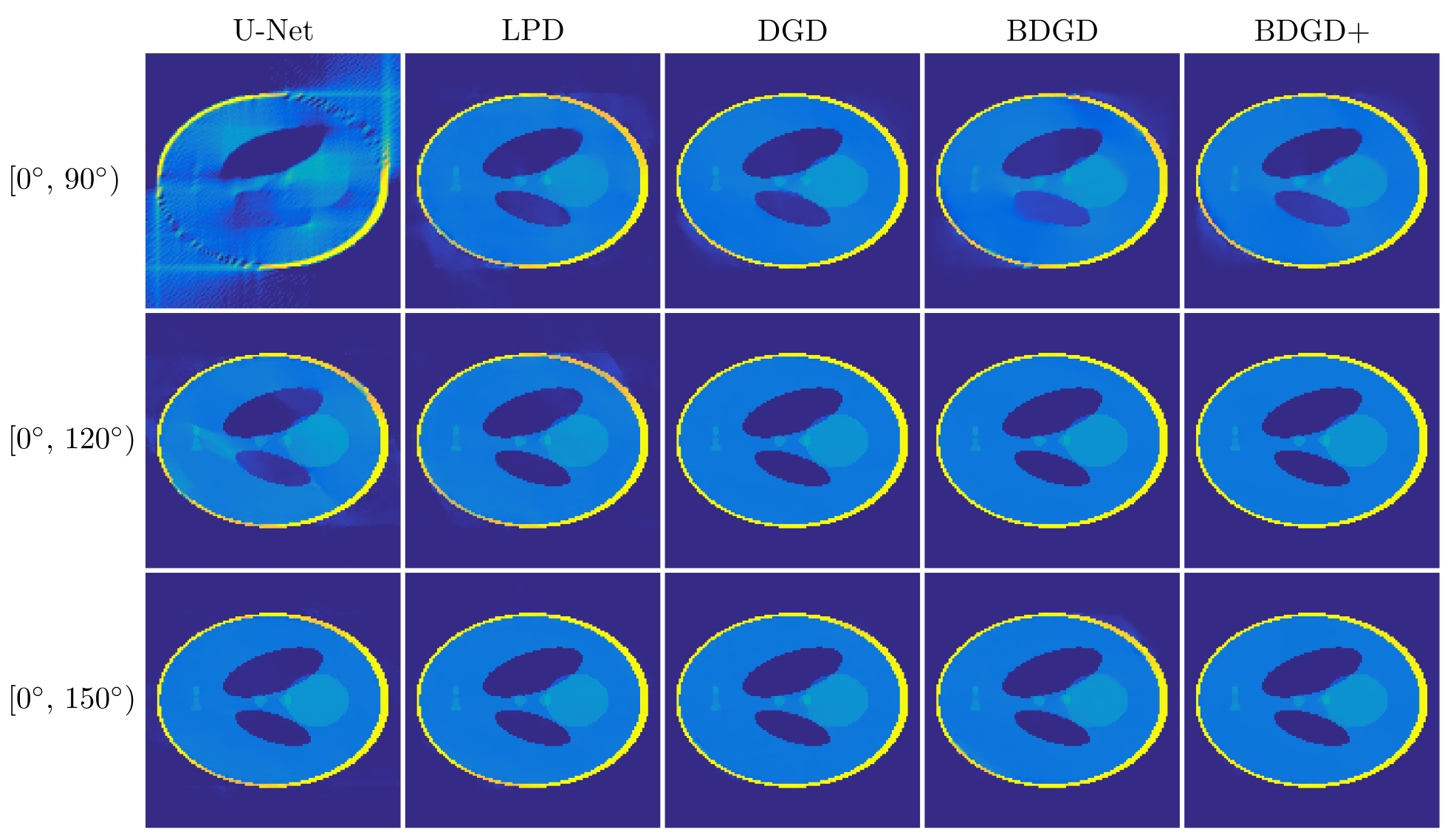}
    \caption{Deep learning approaches for limited angle CT.}
    \label{fig:DeepLimitedView}
\end{figure}

In Fig. \ref{fig:SparseViewMain} we compare how the estimation of uncertainty, in the sparse view case, changes as we add out of distribution text. We also include reconstructed SL phantoms with deep learning-based methods for sparse view and limited angle problems in Fig. \ref{fig:DeepSparseView}, and Fig. \ref{fig:DeepLimitedView}, respectively.

\section{Additional Details on the Architectures}\label{appx:AdditionalArchFigures}

\begin{figure}[ht]
    \centering
    \includegraphics[width=0.9\textwidth]{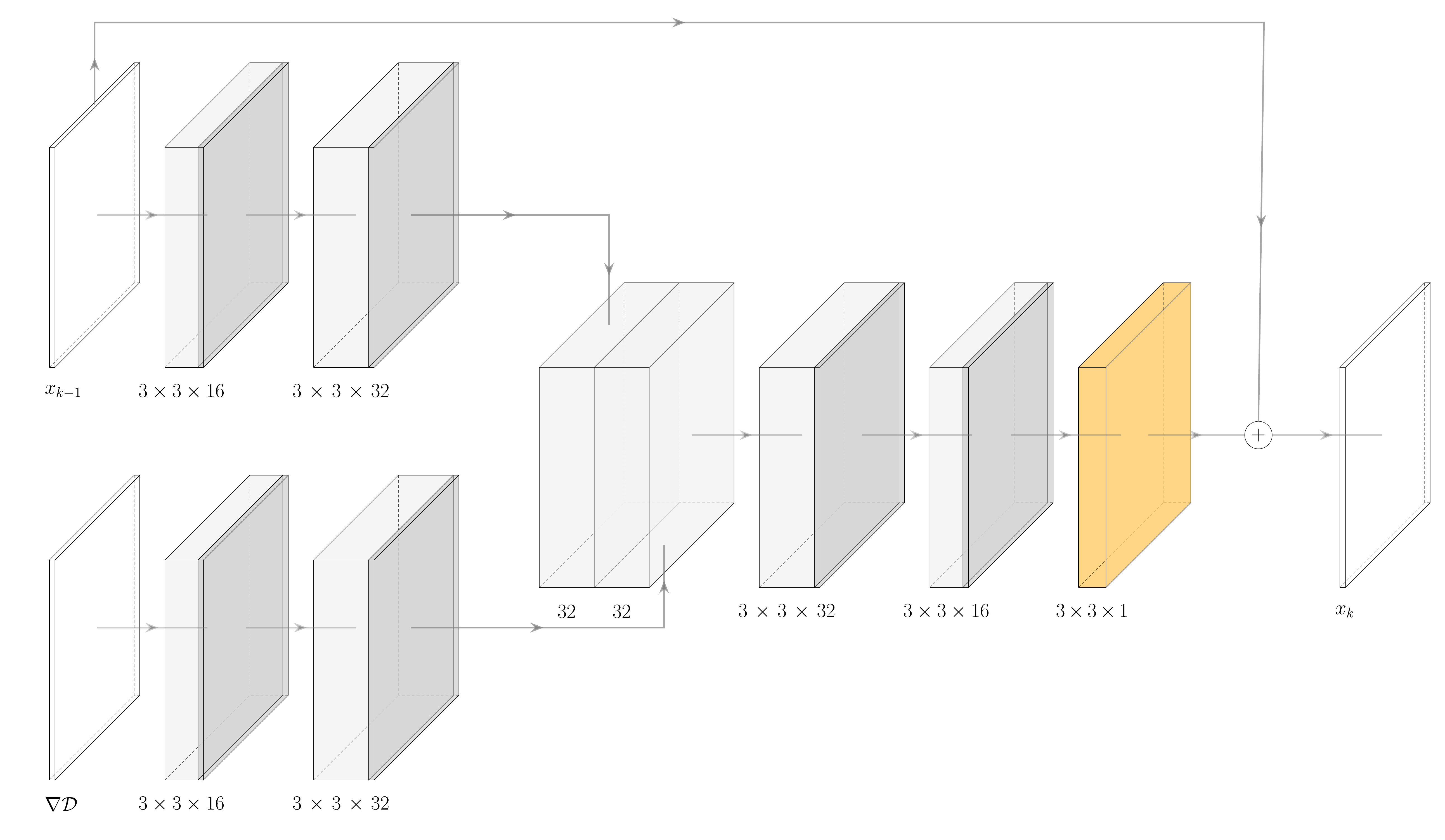}
    \caption{Diagram of the CNN architecture representing one step update with the $k^{\rm th}$ block. The deterministic layers $(\phi_k)$ are coloured grey; the Gaussian mean-field CNN $(\theta_k)$ is coloured yellow.}
    \label{fig:MeanNetSpecs}
\end{figure}

\begin{figure}[ht]
    \centering
    \includegraphics[width=0.6\textwidth]{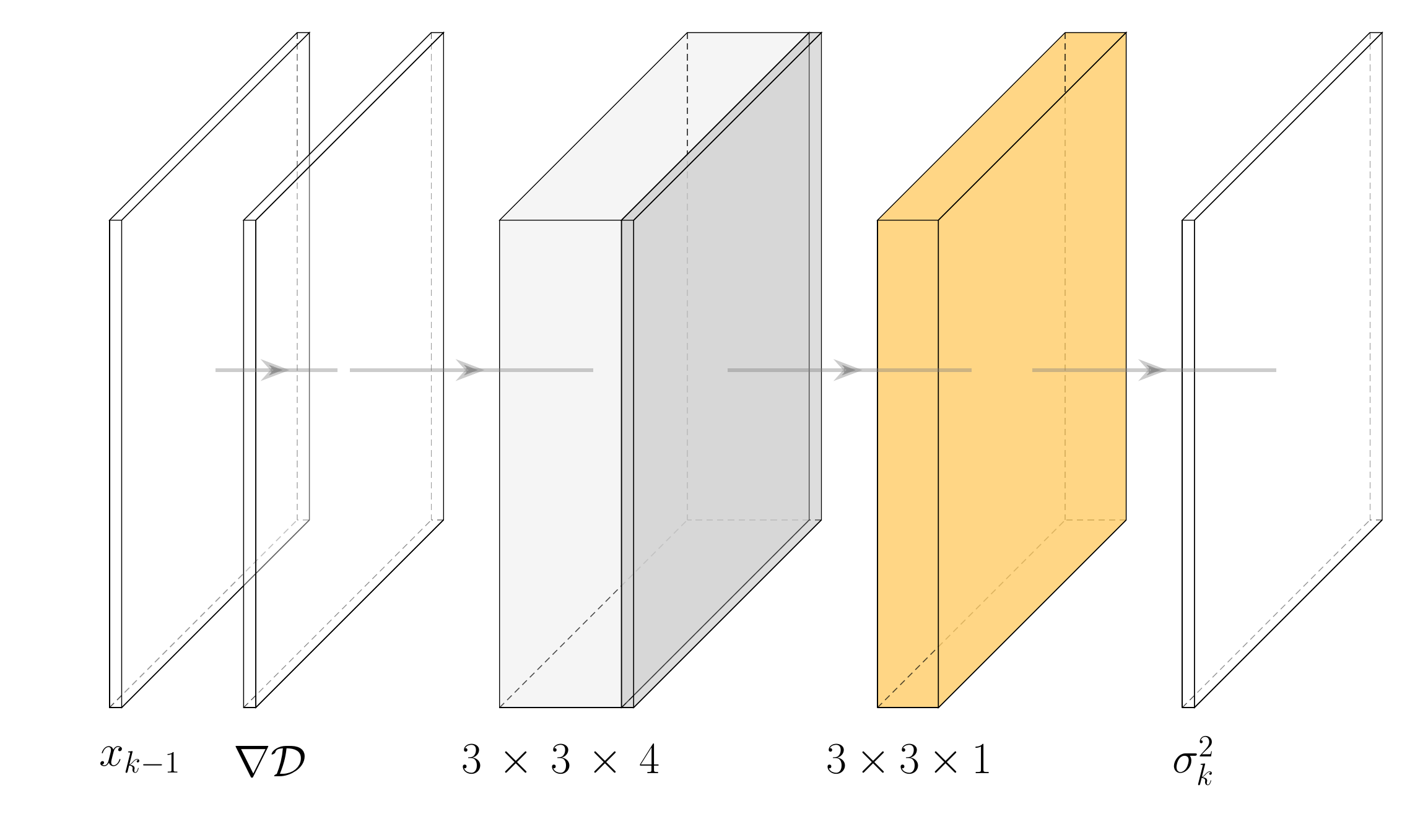}
    \caption{Diagram of the CNN architecture modelling the input-dependent variance with the $k^{\rm th}$ block. The deterministic layers $(\phi_k)$ are coloured grey; the Gaussian mean-field convolutional layer $(\theta_k)$ is coloured yellow.}
    \label{fig:VarianceNetSpecs}
\end{figure}

In Fig. \ref{fig:MeanNetSpecs} we report the architecture of the network representing one step update with the $k^{\rm th}$ block. Fig. \ref{fig:VarianceNetSpecs} shows the architecture used for modelling the input-dependent variance in BDGD+.

\end{document}